# Exploring the weather impact on bike sharing usage through a clustering analysis

Jessica Quach
*Department of Computer Science and Media Technology, Malmö University, Malmö, 205 06, Sweden*
jessica_quach60@hotmail.com

Reza Malekian
*Department of Computer Science and Media Technology, Malmö University, Malmö, 205 06, Sweden*
reza.malekian@mau.se

***Abstract*—Bike sharing systems (BSS) have been a popular traveling service for years and are used worldwide. It is attractive for cities and users who wants to promote healthier lifestyles; to reduce air pollution and greenhouse gas emission as well as improve traffic. One major challenge to docked bike sharing system is redistributing bikes and balancing dock stations. Some studies propose models that can help forecasting bike usage; strategies for rebalancing bike distribution; establish patterns or how to identify patterns. Other studies propose to extend the approach by including weather data. This study aims to extend upon these proposals and opportunities to explore how and in what magnitude weather impacts bike usage. Bike usage data and weather data are gathered for the city of Washington D.C. and are analyzed using k-means clustering algorithm. K-means managed to identify three clusters that correspond to bike usage depending on weather conditions. The results show that the weather impact on bike usage was noticeable between clusters. It showed that temperature followed by precipitation weighted the most, out of five weather variables.**

## I. Introduction

Bicycling has proven to be beneficial in many ways. By traveling by bike instead of motor vehicles, local environmental issues like air pollution and car traffic are mitigated [1]. In the bigger picture, transportation alternatives such as walking and bicycling contributes to a reduction of greenhouse gas emissions [1] and fuel consumption [1], [2], [3]. Research also shows that bicycling encourages a healthy lifestyle because it involves physical exercise [1]. It is a convenient collective transportation for people who wish to contribute to a wholesome nature and live a healthier lifestyle.

Like any other large-scaled systems, issues arise with the implementation of a bike sharing system. Although shared bicycling promotes healthiness for the individual and reduces air pollution; shown in a study by Maizlish et al. [1], there is an increase in traffic accidents with increased bike activities. The authors believe the reason is that bike cities lack adequate infrastructure such as separate bike lanes. Another issue is the redistribution of bikes in docked sharing systems. The bike usage in terms of bike pickup and drop-off differ in each station, which can lead to bike shortages or surpluses in a station. A third issue is the maintenance or condition of the bikes and rebalancing of stations.

The usage of bike sharing systems depends on many factors. All of the aforementioned issues have been pointed out and addressed by multiple studies [4] – [9], but research suggests that there are other factors which need to be examined as well. Weather is a significant factor for bike sharing [7] which is suggested to be examined [9], [4]. In theory, bad weather results in less bike usage [10], [11]. By studying what kind of weather and how much it affects bike usage, this paper could provide useful insight for decision making such as forecasting demand for rebalancing bikes; the number of bikes in a typical location with known seasonal weather would be needed.

The purpose of this paper is to gain a better understanding on how the weather impacts bike sharing usage and to provide empirical data. This information can be useful in determining number of bikes a location would need depending on seasonal weather. In this paper, we will investigate the weather impact on bike usage through a k-means clustering analysis.

## II. Related work

Multiple researches [9], [12], [8], [7], [5] use k-means in their analysis to investigate and explore some patterns with BSS. In their research, Zhao et al. [8] used different clustering algorithms to analyze the characteristics of the spatial distribution of shared traffic resources by investigating the relationship between the bike distribution density and geographical location. Similarly, Ma et al. study [9] used k-means algorithm, hierarchical algorithm and expectation maximization algorithm to explore spatiotemporal activity patterns of a bike sharing system in Ningbo, China. In their study conducted in 2018, Xue and Li [12] used two clustering methods, DBSCAN and k-means, to analyze the traffic patterns in different geographical areas based on a New York City bike sharing system data.

A recent study [9] from 2019 had a similar objective to explore the spatiotemporal activities pattern of bike sharing systems. Using clustering analysis on bike sharing data from Ningbo in China, their results also show that clustered stations have a unique spatiotemporal activities pattern influenced by travel habits and land use characteristics around the stations.

One recurring phenomenon, identified in Xue and Li's paper as well as other studies [12], [13], [9], [7], is the "double-peak" phenomenon for the frequency of bike renting and returning in docking stations. During weekdays, rentals and returns of bikes are drastically increased during traffic rush hours, specifically around eight in the morning and six in the evening. During non-working days, this phenomenon is replaced with a steady curve, increasing in the morning until the afternoon and then decreases as it becomes evening.

Multiple studies [9], [4], [14] have mentioned or suggested weather as a potential factor to study in works related to bike sharing and clustering techniques. As suggested by Ma et al. [9], adding weather as a factor in clustering could uncover new patterns and insights to bike users' behavior when analyzed together with weather data. Liu and George [15] theorized that data mining and examining weather data using fuzzy c-means clustering could give insight and practical benefits. In their research, they managed to use aforementioned techniques to identify spatiotemporal patterns in their data.

In their paper, Chardon et al. [4] mentions that bike sharing usage can vary by season, weather and day of the week. Since inclement weather and weekends means less demand for rebalancing bikes it can be assumed that bike usage was less. When it is a good season and weather, valet services are required in order to prevent stations from overcrowding. Valet services are personnel whose task is to park bikes. Although the paper shows that weather is a significant factor, the authors did not take weather into consideration in their final conclusion.

A study focusing on bike sharing demand in Toronto Canada [11] showed correlations between temperature, land usage and bike usage. In another study, it was revealed that a certain high temperature would reduce bike usage [10]. Precipitation also affected bike usage according to Gebhart and Noland [16]. This shows that different variables that define weather could be further explored and its weight can be investigated.

In conclusion, findings from previous studies show that the usage of shared bikes is highly influenced by geographical and temporal factors such as hour of the day, day of the week and what type of activities take place nearby or in the area. This knowledge will be helpful for preparing data and analyzing the outcomes of this study.

## III. METHOD

### A. Datasets

For this study, two datasets with two years' worth of data have been retrieved. The bike sharing data retrieved from Capital Bikeshare [17], a popular bike sharing system in Washington, D.C., contains historical records of trips in the time period between the 1st of January in 2018 to 31st December 2019. It consists of a total of 694,110,1 observations and 9 features. The data has been processed by Capital Bikeshare to exclude trips that are taken by staff for maintenance and inspection purposes, as well as any trips that are taken for Capital Bikeshare's testing stations at their warehouses. Trips lasting less than 60 seconds are regarded as false starts, or users who are trying to ensure that the bike is securely docked by re-docking it and is therefore excluded.

The weather data, retrieved from Visual Crossing Weather [18], contains historical weather information of Washington D.C. from the time period between the 1st of January in 2018 to 31st December 2019. The data consists of a total of 730 observations and 16 features. Some of the information included in this data are timestamps, average temperature, wind speed, wind gust, precipitation and many more.

### B. Data Preparation

The main focus in this section is to describe how the data are cleaned, reduced or transformed, and to present the resulting dataset. As seen in previous work, the bike usage is highly influenced by the hour of the day. In order to avoid bias from the hourly bike usage such as the "double peak" phenomenon, this study will focus on the daily basis relations rather than hourly basis.

#### 1) The Bike Sharing Dataset

All data of the bike sharing dataset in the project are treated according to the EU law: General Data Protection Regulation (GDPR) 2016/679. For this research, the only interesting feature of the bike dataset is the number of trips that were taken each day. Thus, data such as station number, bike number and membership are ignored and removed from the dataset. The number of trips taken each day is summarized into a new feature "*Count*".

#### 2) The Weather Dataset

The secondary dataset contains features such as "Address" and "Location" which is removed because the weather information covers the area of Washington, D.C. The features "Snow Depth", "Wind Chill" and "Heat Index" contain more than 50% of missing data or no data at all and are removed from the dataset, as the existing data is insufficient to calculate any valid approximations to replace the missing values [19].

"Wind Gust", containing 29.7% missing values, is removed from the dataset after a consultation with Visual Crossing Technical Support [20]. As it is written in the documentation [18], wind gust is only recorded by weather stations if the measured short-term wind speed to be significantly more than mean wind speed. If the wind gust does not meet the criteria, a null or empty value is returned. However, as said in the documentation, these values do not indicate that there were no wind gusts at all.

The feature "Conditions", or weather types, can hold any of the values: "clear", "overcast", "partially cloudy", "rain, clear", "rain, overcast" or "rain, partially cloudy". The values of this feature are nominal and cannot be measured in any way. Clustering algorithms cannot handle nominal data and would have to be converted to numeric values. For example, let the value "clear" have the numeric value 0, "overcast" the value 1, "rain" the value 2 and so on. However, this leads to the issue that the k-means will treat the value represented by 0 to be closer to the value represented by 1 than 2. This does not make sense in the real-world, as there is no way to measure the closeness between said weather conditions. Furthermore, the values of this feature can be accounted for by the features "Cloud Cover", "Precipitation" and "Relative Humidity". The feature "Conditions" is removed for these reasons.

#### 3) Aggregation of Datasets

Once both datasets have been processed and cleaned individually, they are combined into one dataset. Since we are interested in which of the features have the most impact on bike usage, we look at the correlation coefficients between the features in the dataset which can be seen in Figure 1. Features which have a strong correlation with each other except for correlation with bike usage may be removed in order to increase performance of the k-means algorithm. In the figure, 1.0 indicates a strong positive correlation between two features, 0 indicates no correlation between two features and -1.0 indicates a strong negative correlation between two features.

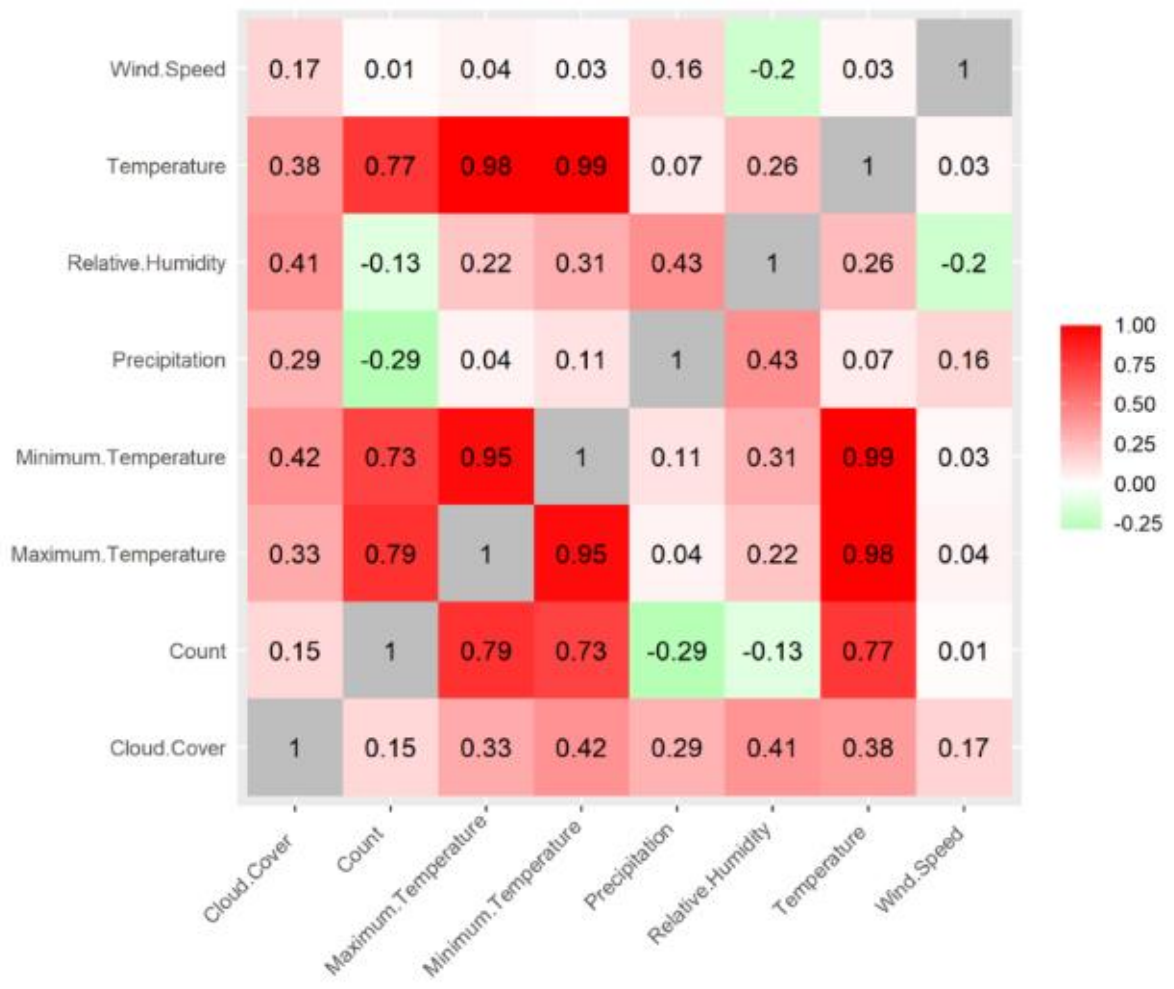

*Figure 1. A correlation matrix for visualizing the correlation between features in the dataset.*

Figure 1 is a correlation matrix which shows that "Temperature", "Maximum Temperature" and "Minimum Temperature" are strongly correlated. They also have a strong correlation with the number of trips "Count". We remove the features for maximum and minimum temperature and keep the average temperature "Temperature". The feature "Temperature" seems to have the highest correlation with "Count", followed by the feature "Precipitation". The strong positive correlation between trips and temperature indicate that the number of trips increases as the temperature increases. The negative correlation between trips and precipitation indicate that number of trips decrease as the precipitation increase.

The result of the data preparation is a dataset which includes 668 observations and 6 features, where each observation hold record for a day. The features are presented and briefly explained in the following list:

- "Count": integer number, number of trips.
- "Temperature": decimal number, average temperature measured in Fahrenheit.
- "Precipitation": decimal number, average rainfall measured in inches.
- "Wind Speed": decimal number, average wind speed measured in miles per hour.
- "Cloud Cover": decimal number, average cloud cover measured in percent.
- "Relative Humidity": decimal number, average humidity measured in percent.

Features with the necessity to be standardized are scaled to have the standard deviation one. This removes possible dissimilarity caused by the features' unit measurement [21]. The dataset has been cleaned and processed into one dataset and is now ready to be used for the k-means algorithm.

### C. Modelling

In this section, the chosen cluster validation methods and k-means clustering algorithm will be presented in detail.

#### 1) Cluster Validation

Cluster validation methods are used to determine the optimal number of clusters k to use when computing the k-means algorithm. These methods, as well as the k-means algorithm, must use a distance metric to compute the distance between two data points. For this task, the Euclidean distance has been chosen and the formula is given below:

$$D = \sqrt{\sum_{i=1}^{n} (x_i - y_i)^2} \qquad (1)$$

where $D$ is the calculated distance between two vectors $x$ and $y$ of length $n$.

Three cluster validation methods have been chosen to find k, these are the gap statistic, the silhouette method, and the elbow method.

Method 1: *the gap statistic*. The formula [22] for performing gap statistic is given below:

$$Gap_n(k) = E_n^* \log(W_k) - \log(W_k) \qquad (2)$$

where $Gap_n(k)$ is the estimated gap statistic, $E_n^*$ denotes the expectation under a sample size n from the reference distribution, and $(W_k)$ is the within sum of squares. The estimation of k will be the value which maximizes the gap statistic $Gap_n(k)$ after the sampling distribution has been considered.

Method 2: *the average silhouette width*. The formula [23] for performing silhouette is given below:

$$s(i) = \frac{b(i) - a(i)}{max\{a(i), b(i)\}} \qquad (3)$$

where $a(i)$ is the average dissimilarity of an object $i$ to all other objects in its cluster and $b(i)$ is the average dissimilarity of $i$ to all other objects in the nearest cluster that is not its own nearest cluster defined by the cluster minimizing the average dissimilarity. The estimation of $k$ will be the value which maximized the average silhouette $s(i)$ over a range of possible values for $k$. It is notable that this method assumes that the number of clusters is more than one.

Method 3: *the total within-cluster sum of square (elbow)*. The formula [24] for performing the elbow method is given below:

$$W_k = \sum_{r=1}^{k} \frac{1}{n_r} D_r \qquad (4)$$

where $W_k$ is the within-cluster sum of squares, $k$ is the number of clusters, $n_r$ is the number of points in cluster $r$ and $D_r$ is the sum of distances between all points in a cluster. The estimation of $k$ will be the value which minimizes the within-cluster sum of square $W_k$.

#### 2) Computing k-means

The Hartigan-Wong approach is used to compute k-means. The aim of this approach is to search for a k-partition with locally optimal within-cluster sum of squares by moving data points from one cluster to another [25]. The Hartigan-Wong algorithm recomputes the centroids any time a data point is moved. It also computes differently in order to increase time and accuracy. The maximum number of iterations is set to 10 due to computation limitations and the number of configurations of centroids is set to 25.

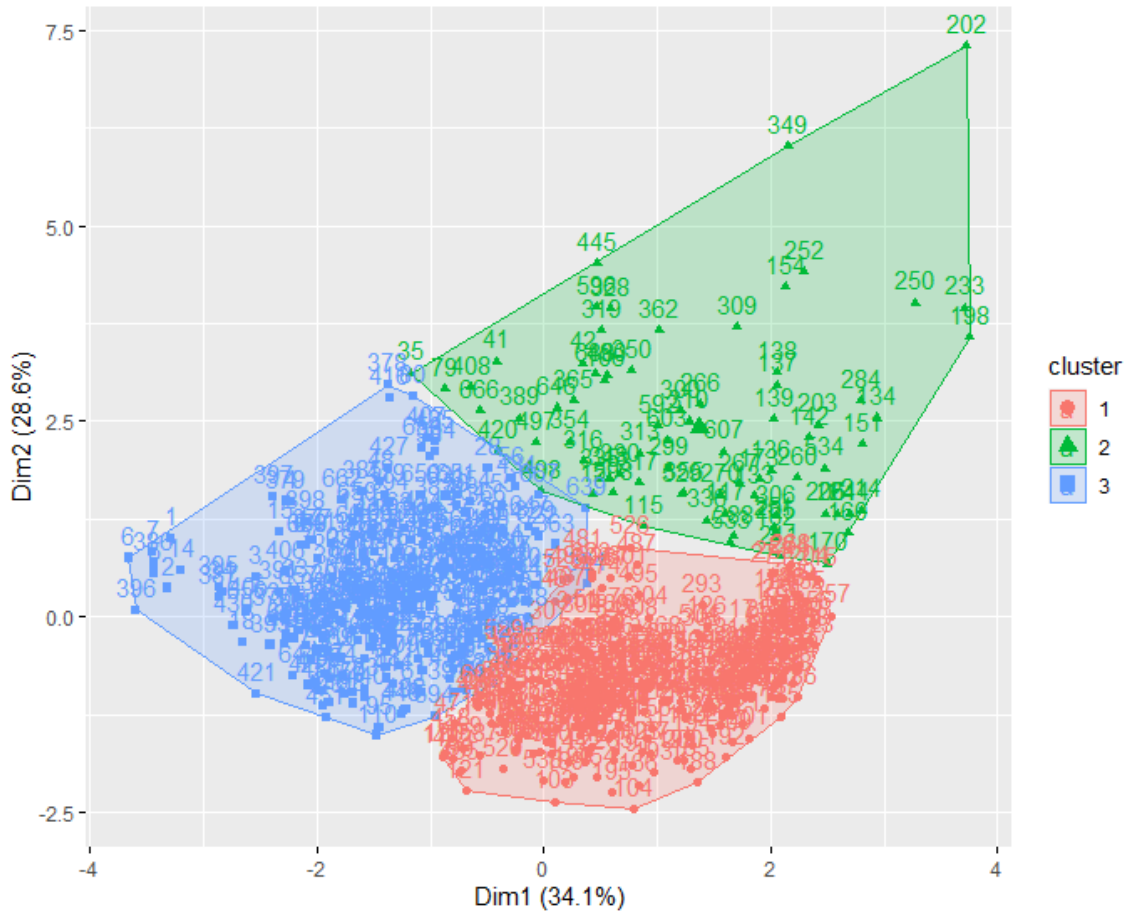

*Figure 2. The visualized result of computing k-means with k=3.*

## IV. RESULTS

In this section, the result from the cluster validation and the k-means algorithm is presented and explained.

### A. *The Result of Cluster Validation*

The results for the cluster validation are presented in the Table 1 below:

*TABLE 1. Results from the cluster validation methods for estimating k.*

| Method: | Elbow method | Silhouette method | Gap statistic |
|---|---|---|---|
| Optimal number of clusters (*k*) | 3 | 3 | 4 |

The elbow and silhouette methods yield the optimal value *k* to be 3 whereas gap statistic yield 4 clusters. Therefore, we choose to set *k* to hold the value of 3.

### B. *The Result of K-means*

The result of clustering the datasets using k-means with 3 clusters is visualized and shown in Figure 2.
The 668 observations are clustered into 3 clusters. A total of 305 observations have been grouped into *Cluster 1* which is marked red, 78 observations in *Cluster 2* which is marked green, and 285 observations in *Cluster 3* which is marked blue.

The first dimension on the x-axis explains 34.1% of the variance in the data, and the second dimension explains 28.6%. This means that this visualization is able to capture 62.7 % of the variance in the data. The two dimensions are called principal components which captures as much of the information in the data as possible. Details about each cluster's minimum, maximum, median and mean value for each feature is presented in Table 2 – Table 4 below. The values have been rounded up to have 2 decimals. The features' label has been shortened to "Temp." for "Temperature", "Preci." for "Precipitation", "WS" for "Wind Speed", "CC" for "Cloud Cover" and "RH" for "Relative Humidity".

*TABLE 2. Detailed information about Cluster 1, marked red in Figure 2.*

| | **Count** | **Temp.** | **Preci.** | **WS** | **CC** | **RH** |
|---|---|---|---|---|---|---|
| Min. | 6183 | 48.40 | 0.00 | 4.90 | 0.30 | 32.32 |
| Median | 12 581 | 73.80 | 0.00 | 12.70 | 30.00 | 65.24 |
| Mean | 12 647 | 71.83 | 0.03 | 12.86 | 43.93 | 64.82 |
| Max. | 19 113 | 88.00 | 0.60 | 28.10 | 100.0 | 88.53 |

*TABLE 3. Detailed information about Cluster 2, marked green in Figure 2.*

| | **Count** | **Temp.** | **Preci.** | **WS** | **CC** | **RH** |
|---|---|---|---|---|---|---|
| Min. | 1208 | 34.20 | 0.10 | 5.70 | 17.10 | 70.54 |
| Median | 6530 | 61.55 | 0.85 | 14.10 | 91.20 | 83.84 |
| Mean | 6363 | 60.21 | 0.98 | 14.43 | 72.98 | 83.53 |
| Max. | 13 709 | 82.70 | 4.00 | 25.40 | 100.00 | 99.10 |

*TABLE 4. Detailed information about Cluster 3, marked blue in Figure 2.*

| | **Count** | **Temp.** | **Preci.** | **WS** | **CC** | **RH** |
|---|---|---|---|---|---|---|
| Min. | 628 | 14.70 | 0.00 | 3.40 | 0.30 | 20.36 |
| Median | 6829 | 42.20 | 0.00 | 11.70 | 18.40 | 57.40 |
| Mean | 6564 | 41.46 | 0.05 | 12.88 | 23.75 | 58.93 |
| Max. | 11 740 | 58.50 | 0.70 | 33.00 | 100.00 | 97.34 |

#### 1) *Cluster Comparision: Bike Usage Frequency*

The tables for each cluster show dissimilarities between the clusters. In Table 2, we can see that bike usage frequency varies between 6183 to 19113 trips a day for *Cluster 1*. This is significantly higher than in *Cluster 2* and in *Cluster 3;* the variance is between 1208 to 13709, and 628 to 11740 respectively. Another significant gap between *Cluster 1* and the two other clusters can be seen in the average bike usage frequency for all observations in each cluster. *Cluster 2* and *Cluster 3* show a mean value of 6363 trips per day and 6564 respectively. This is only half of the average bike usage of *Cluster 1*, which has the mean value of 12647 trips per day. To get a better understanding of the clustering, we look at the weather conditions for each cluster and compare it with the other clusters.

#### 2) *Temperature Impact on Bike Usage*

We begin with temperature, as this feature was most correlated with bike usage frequency according to the correlation matrix presented in Figure 1. As seen in the tables, both the mean temperature and the median temperature is highest in *Cluster 1*, followed by *Cluster 2* and lastly *Cluster 3*. The range between the minimum temperature and maximum temperature is also highest in *Cluster 1*; with the minimum of 48.40 degrees Fahrenheit and the maximum 88.00 degrees Fahrenheit. In *Cluster 2*, the range is slightly lower and varies between 34.20 degrees Fahrenheit to 82.70 degrees Fahrenheit. The minimum value for temperature in *Cluster 3* is as low as 14.70 degrees Fahrenheit and the maximum value is 58.50 degrees Fahrenheit. This shows that trips within *Cluster 1* were done during warmer temperatures; trips within *Cluster 2* were done during temperatures slightly lower than within *Cluster 1*, and trips within *Cluster 3* were done during low temperatures. This also shows that the temperature certainly impacts the bike usage, the higher the temperature, the higher the bike usage. The opposite is also true; the lower the temperature, the lower the bike usage.

#### 3) *Precipitation Impact on Bike Usage*

Precipitation is measured in inches and is the second feature that correlated highly with bike usage frequency. We can see that the mean precipitation for *Cluster 1* is as low as 0.03 inches and that it rained 0.60 inches at most. The range between minimum and maximum precipitation for *Cluster 2* is 0.10 inches to 4.00 inches. The median and average precipitation are 0.85 inches and 0.98 inches respectively. As for *Cluster 3*, the mean is 0.05 inches and the highest is 0.70 inches. This shows that trips within *Cluster 1* took place when there was very little or no precipitation at all. The same is true for *Cluster 3* which differentiates faintly from *Cluster 1*. We can see a somewhat significant gap between *Cluster 2* and the other two clusters; trips within *Cluster 2* was done during precipitation compared to the others. To summarize the interpretation of precipitation impact on bike usage; the higher the precipitation, the lower the bike usage. Conversely, the lower the precipitation, the higher the bike usage.

#### *4) Cloud Cover, Humidity and Wind Speed*

As we can see in the correlation matrix, Figure 1, cloud cover has a small correlation with bike usage frequency. This is revealed in the results. In *Cluster 2*, the mean and median cloud cover is above 70% while it is lower than 50% for *Cluster 1*, and lower than 25% for *Cluster 3*. This shows that trips within *Cluster 1* and *Cluster 3* were done when the weather was sunny, or partly cloudy. In *Cluster 2*, most trips were done when it was cloudier. However, this cannot indicate that higher cloud cover percentage equals higher bike usage frequency. Looking at *Cluster 1*, which has many more trips taken, the overall cloud cover is higher than in *Cluster 3*. This is because of the strong correlation and impact temperature has on the bike usage.

The average and mean humidity for each cluster seems to differ by a small margin. However, it is observable that the humidity percentage is lower in *Cluster 1* and *Cluster 3* compared to *Cluster 2*. Like cloud cover, humidity has a weak impact on bike usage frequency.

Wind speed has an insignificant correlation with bike usage frequency. This is reflected in the tables for each cluster; we can see that there is no certain pattern.

To summarize, we can see from the results that the bike usage frequency is highest in *Cluster 1*. In this cluster, the temperature is high and there is little or no precipitation. In *Cluster 2*, the bike usage is half as frequent than in *Cluster 1*. Moreover, the temperature is somewhat lower and there is more precipitation than in *Cluster 1*. In *Cluster 3*, the bike usage frequency is lower than *Cluster 1* and possibly also lower than in *Cluster 2*. What is notable in *Cluster 3* is that, although there is little or no precipitation, the bike usage is still lower than in *Cluster 1*. This is because of the low temperature in *Cluster 3*.

## V. ANALYSIS

In this section, an analysis on the results will be presented and interpretation of the results will be discussed.

### *A. Anomaly Analysis*

While the dataset was processed to eliminate possible outliers such as data-entry errors or measurement errors, the results seen in Figure 2 show that are a few observations which may seem to be anomalies. It may be worthwhile to take into consideration and investigate unusual events or phenomenon that may cause abnormal results in the data. Global warming and climate change increase the probability of weather anomalies [26]. Weather and climate changes cause extreme weather events such as wildfires and heavy rain to occur more often. In this section, an analysis is done on unusual observations within the dataset.

The observation to the upper-right of the figure 2 is labeled as 202 and is clustered into *Cluster 2*. This observation will be referred to as *Observation 202*. The observation that is somewhat close to *Observation 202*, labeled as 349, also seem to diverge from the rest of the observations and is also clustered into *Cluster 2*. This observation will be referred to as *Observation 349*.

*Observation 202* was the 21st of July in 2018. It was a Saturday with the average temperature at 71.7 degrees Fahrenheit, precipitation at 4.0 inches and wind speed at 21.4 miles per hour. The number of trips made that day was 2794 and this day contained the heaviest rainfall in the whole dataset. According to The Washington Post [27], there was a storm this day that ranked as the fifth wettest July day of Washington D.C. The heavy rain brought by the storm lead to flooding and strong winds which is most likely the cause of the low bike usage frequency.

*Observation 349* was Saturday 15th of December in 2018. The average temperature was 51.5 degrees Fahrenheit, precipitation was 2.5 inches and wind speed at 18.4 miles per hour and a total of 1208 trips were recorded this day. Although there were no unusual events or occurrences, the high precipitation alone may be the reason to why the bike usage was low.

### *B. Analysis of Working Days and Non-Working Days*

In this section, we analyze bike usage difference between working days and non-working days. The calendar from QuantLib [28] is used for distinguishing working days from non-working days. Days included in the non-working days group are Saturdays, Sundays, and public holidays such as New Year's Eve and Christmas. All other days are included in the working days group. It is presented in Figure 3.

Looking at Figure 3, there are generally higher bike usage in each cluster during working days than the corresponding cluster during non-working days; this shows that there may still be some bias involving day of week. However, it can be seen in the figure that the bike usage frequency for *Cluster 1* does not differentiate much for the day of week.

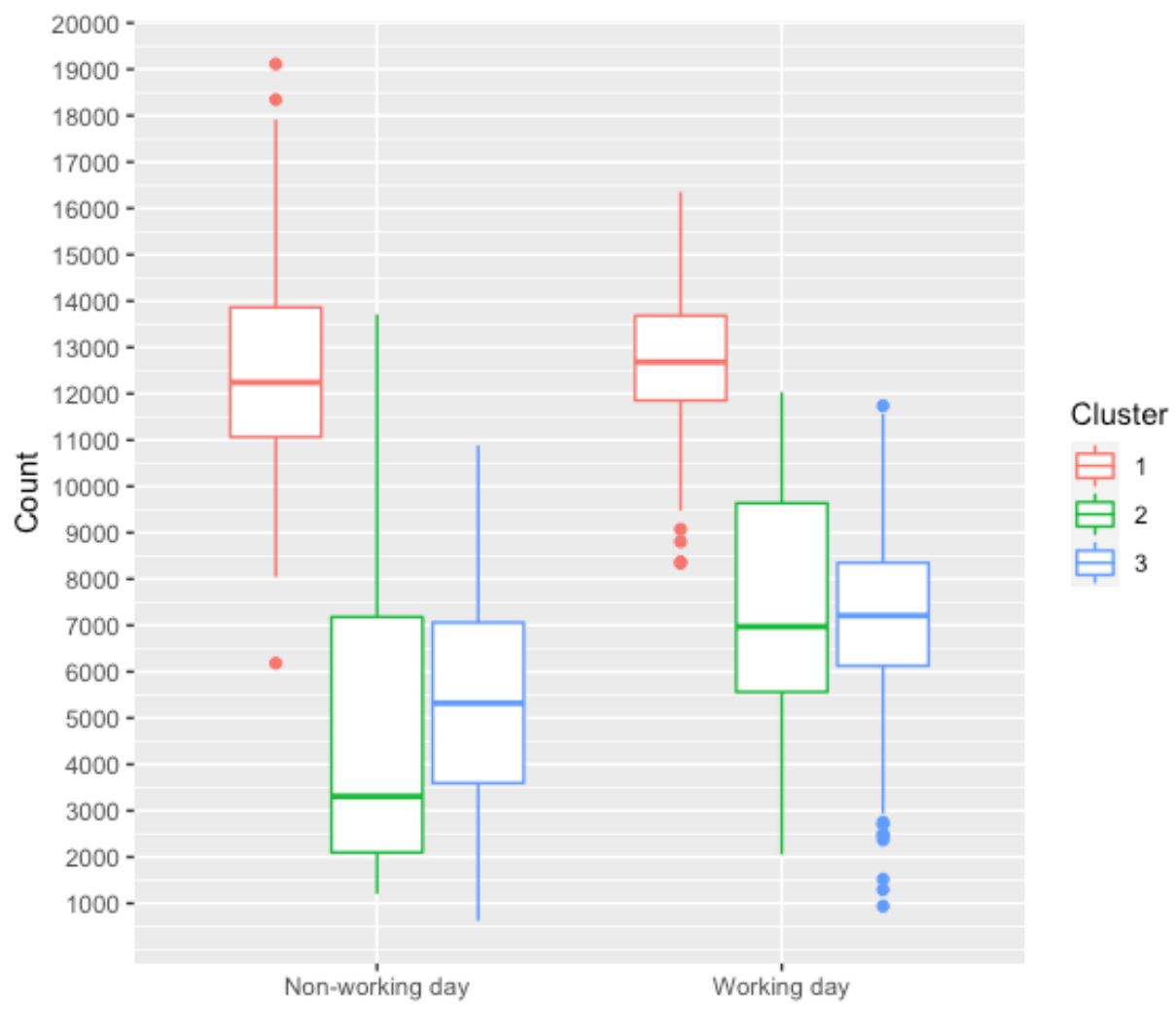

*Figure 3. Bike usage in working days and non-working days for each cluster.*

Public transportation services are a common travelling method for going to and from work or school. It is also useful for running errands such as grocery shopping or meeting up with people. The double-peak phenomenon observed from other studies presented in related works shows that there is usually higher bike usage frequency in the morning and evening in working days. As a result, there may be more bicycle trips made in working days. The studies also show that the weekends and holidays have a steadier curve for use of bike services, mostly for running errands or other activities between noon and evening. However, the majority of observations in *Cluster 1* for working days and non-working days seem to have about the same amount of bike usage. *Cluster 1* contained observations with the highest temperature and lowest precipitation. This cluster may indicate that while day of week may matter, the weather conditions have a deciding impact on whether people decide to bike or not. As for the case for trips in *Cluster 1*, the warm temperature and absence of precipitation may have resulted in the high amount of bike usage.

### *C. Seasonality Analysis*

By analyzing the bike usage frequency based on seasonality, it may be possible to uncover additional patterns. This analysis will show bike usage for each season in each cluster. It will also show bike usage for each season without the clusters.

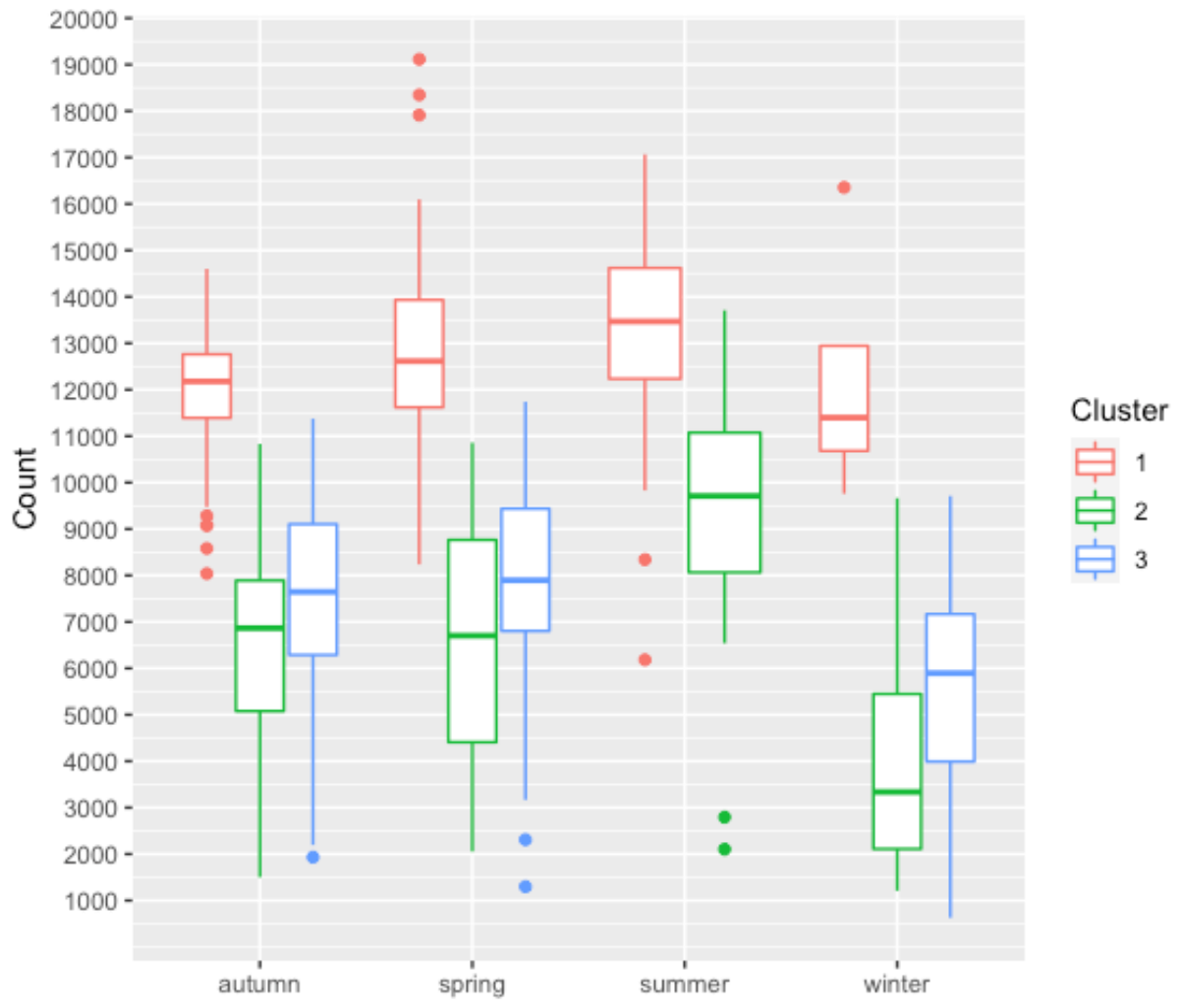

*Figure 4. Bike usage for each season in each cluster.*

The figure above shows bike usage frequency (Count) for each season in the clusters. One observation that can be made from the figure is that *Cluster 3*, which contained trips made in colder temperatures, do not contain any trips made during summer. Another observation is that bike usage frequency for each cluster is lowest during winter comparing to the corresponding clusters in the other seasons. Generally, it is common that the temperature is lower during winter and higher during summer. This may have resulted in clusters that depend on temperature which in turn depend on seasonality. Therefore, an analysis without the clusters is made.

Looking at the dataset: there are 182 observations made in autumn with a total of 1,792,486 trips; 184 observations with a total of 1,881,527 trips made in spring; 122 observations with a total of 1,551,109 trips made in summer; and 180 observations with a total of 999,290 trips made in winter. Since there are an uneven amount of observations for each season, we calculate the average number of trips made per day in each season to see which season contained the most or least trips. This is done by dividing the total number of trips for a season with the number of observations for the season. The equation can be seen below:

$$Average = \frac{total\ number\ of\ trips_s}{observations_s} \quad (5)$$

where $s$ is the season.

The average bike trips per day for autumn, spring, summer, and winter are 9849, 10226, 12714 and 5552, respectively. This shows that bike trips are highest during summer followed by spring and autumn. The number of bike trips during winter is considerably lower.

## VI. CONCLUSION AND FUTURE WORK

The conclusion drawn from this paper's results and analysis is that weather has a significant impact on bike usage. Specifically, temperature weighted the most, followed by precipitation, out of five weather variables. High temperature and low or no precipitation lead to high bike usage frequency; high temperature and high precipitation lead to low bike usage frequency; low temperature and low or no precipitation lead to low bike usage frequency. Noted in the analysis, there may be traces of bias and limitations to this study. Despite that, it does not change the weather pattern and impact found regarding bike usage frequency. Some suggestions for future work are to increase the amount of data used and to include more variety by investigating multiple BSS in different cities. In addition, it might be interesting to examine the impact of trip distance or working days and non-working days. The results and knowledge derived from this paper contributes to empirical data on how the weather impacts the use of BSS. Furthermore, this can help cities and travel agencies in planning and decision-making regarding implementation of BSS.

## ACKNOWLEDGMENT

The authors of this paper would like to thank Visual Crossing for their consultation regarding weather and Nils Lundahl for the time he spent on analysis support and suggesting improvements. The authors would also like to acknowledge the use of datasets from Capital Bikeshare and Visual Crossing Weather.